\documentclass[runningheads]{jbds}
\usepackage[section]{placeins}
\usepackage[numberedbib]{apacite} %% APA style
\usepackage{listings} %% include programming lists
\usepackage{graphicx}
\usepackage{xspace} 
\usepackage[british]{babel}
\usepackage[utf8]{inputenc}
\usepackage{csquotes}
\usepackage[flushleft]{threeparttable}
\usepackage{multirow}
\usepackage{authblk}
\usepackage{longtable} 
\usepackage{url}
\usepackage{changes}
\usepackage{float}
\usepackage{xcolor}
\usepackage[toc, page]{appendix}
\usepackage{lscape}
\usepackage{afterpage}
\usepackage{esvect}
\usepackage{amsmath}
\usepackage{ragged2e}
\justifying\let\raggedright\justifying
\usepackage{enumitem}
\usepackage{makecell}
\usepackage[nodisplayskipstretch]{setspace}
\usepackage{subcaption} 
\usepackage{adjustbox}

\usepackage{array}
\newcommand{\PreserveBackslash}[1]{\let\temp=\\#1\let\\=\temp}
\newcolumntype{C}[1]{>{\PreserveBackslash\centering}p{#1}}
\newcolumntype{R}[1]{>{\PreserveBackslash\raggedleft}p{#1}}
\newcolumntype{L}[1]{>{\PreserveBackslash\raggedright}p{#1}}
\begin{document}
% journal information, do not change
%\journalYear{2024}
%\journalVol{1}
%\journalNo{1}
%\startPage{100}
%\journalPage{100--113}
%\paperDOI{10.35566/jbdsv1n1p1}
% title of the article
\title{Machine Learning Approaches for Mental Illness Detection on Social Media: A Systematic Review of Biases and Methodological Challenges}
\titlerunning{Systematic Review on Mental Illness Detection} % If the paper title is too long for the running head, 
% you can set an abbreviated paper title here as the running title

% authors of the paper. Multiple authors separated by \and
% \inst{1} gives the institute number
\author{Yuchen Cao\inst{1,}\thanks{These authors contributed equally to this work.} \and
Jianglai Dai\inst{2,*} \and
Zhongyan Wang\inst{3} \and
Yeyubei Zhang\inst{4} \and
Xiaorui Shen\inst{1} \and
Yunchong Liu\inst{4} \and
Yexin Tian\inst{5} \thanks{Corresponding author}}
\authorrunning{Y. Cao, J. Dai, et al.}
% First names are abbreviated in the running head.
% If there are more than two authors, 'et al.' is used.

\institute{
Khoury College of Computer Science, Northeastern University, USA \and
Department of EECS, University of California, Berkeley, USA \and
Center for Data Science, New York University, USA \and
School of Engineering and Applied Science, University of Pennsylvania, USA \and
Georgia Institute of Technology, College of Computing, USA
}

\maketitle  % typeset the header of the contribution
\begin{abstract}
The global increase in mental illness requires innovative detection methods for early intervention. Social media provides a valuable platform to identify 
mental illness through user-generated content. This systematic review examines machine learning (ML) models for detecting mental illness, with a particular focus on depression, using social media data. It highlights biases and methodological challenges encountered throughout the ML lifecycle. A search of PubMed, IEEE Xplore, and Google Scholar identified 47 relevant studies published after 2010. The Prediction model Risk Of Bias ASsessment Tool (PROBAST) was utilized to assess methodological quality and risk of bias. The review reveals significant biases affecting model reliability and generalizability. A predominant reliance on Twitter (63.8\%) and English-language content (over 90\%) limits diversity, with most studies focused on users from the United States and Europe. Non-probability sampling methods (approximately 80\%) limit representativeness. Only 23\% of studies explicitly addressed linguistic nuances like negations, crucial for accurate sentiment analysis. Inconsistent hyperparameter tuning was observed, with only 27.7\% properly tuning models. About 17\% did not adequately partition data into training, validation, and test sets, risking overfitting. While 74.5\% used appropriate evaluation metrics for imbalanced data, others relied on accuracy without addressing class imbalance, potentially skewing results. Reporting transparency varied, often lacking critical methodological details. These findings highlight the need to diversify data sources, standardize preprocessing protocols, ensure consistent model development practices, address class imbalance, and enhance reporting transparency. By overcoming these challenges, future research can develop more robust and generalizable ML models for depression detection on social media, contributing to improved mental health outcomes globally.


\end{abstract}

\setcounter{footnote}{0}

\section{Introduction}
Mental health disorders, including depression, represent a critical global health challenge, impacting approximately 1 in 8 people worldwide—approximately 970 million individuals in 2019 \cite{who2023depression}. Depression, one of the most prevalent mental health conditions, affects over 280 million individuals globally, including around 23 million children and adolescents. The COVID-19 pandemic has further exacerbated mental health issues, with notable increases in depression and anxiety observed during this period \cite{who2023depression}. The prevalence of mental health conditions, especially depression, highlights an urgent need for innovative detection methods and interventions. Early identification can lead to more effective treatment outcomes, alleviating the burdens placed on individuals, their families, and healthcare systems \cite{kessler2017trauma}. 

In today's digital age, social media platforms such as Twitter, Facebook, and Reddit play a central role in daily life for millions of people. Studies have shown that individuals often openly express their thoughts, emotions, and mental states on Twitter, making it a valuable platform for examining mental health trends and developing tools for detection and intervention \cite{dechoudhury2013social}. The extensive user-generated content on these platforms provides a unique opportunity for mental health research, enabling the real-time analysis of linguistic patterns and behavioral trends, and providing insights that may otherwise be inaccessible \cite{guntuku2017detecting}.

Advancements in machine learning and deep learning have significantly enhanced the ability to process and analyze large-scale datasets. These technologies are particularly suited for handling the complex and nuanced data found on social media, as they identify patterns and make predictions based on textual and behavioral cues. This capability offers practical tools for mental health detection, allowing researchers to develop models that can potentially identify at-risk individuals based on their social media activity \cite{shatte2019review}. By leveraging algorithms capable of learning from such diverse and rich datasets, researchers are able to develop models that contribute to early intervention efforts in mental health care.

\subsection{Overview of Historical Studies on Machine Learning Approaches for Mental Health Detection in Social Media}
A growing body of research has explored the application of machine learning techniques to detect depression through social media platforms. Approaches range from traditional machine learning techniques such as logistic regression and support vector machines to advanced deep learning models and ensemble methods---have been employed to classify user posts and predict mental health conditions based on linguistic features, patterns, and metadata \cite{dechoudhury2013social, yazdavar2020multimodal, calvo2017nlp}. Platforms like Twitter, Facebook, and Reddit are frequently utilized due to their large user bases and the accessibility of publicly available text-based data. In contrast, TikTok, with its short-video format, provides a distinct medium that captures audiovisual cues such as tone, facial expressions, and gestures, providing researchers with additional dimensions for understanding mental health dynamics.

One of the most common approaches within this research involves sentiment analysis, which aims to determine the emotional tone of user-generated content. By assessing positive, negative, or neutral sentiment \cite{kumar2020sentiment}, researchers attempt to correlate language patterns with indicators of depression. For instance, several studies have examined the use of first-person singular pronouns and negative emotion words as potential depression signals \cite{rude2004language}. 

Despite promising results, multiple challenges remain. First, many studies suffer from limited generalizability due to small or homogeneous samples that may not represent the broader population. Data bias is a significant concern, stemming from the overrepresentation of certain demographic groups or linguistic communities while underrepresenting others \cite{olteanu2019social}. Moreover, the dispersion of research in advanced machine learning methods for mental health detection across the literature, combined with a lack of robust sampling methods and standardized protocols, impedes the reliability of findings. Additionally, insufficient handling of complex linguistic nuances, such as context-dependent meanings, further limits the effectiveness of these detection efforts \cite{calvo2017nlp}.

\subsection{Research Gaps and Objectives of the Current Study}
While individual studies have provided valuable insights into the application of machine learning for mental health detection, significant gaps persist in the literature. These include the broader implications of biases and limitations across studies and the lack of comprehensive reviews consolidating the effectiveness of machine learning models \cite{calvo2017nlp}. Additionally, existing research does not consistently address methodological challenges across different stages of machine learning applications, such as sampling, preprocessing, model development, and evaluation \cite{thieme2020machine}. Therefore, a systematic review is essential to unify findings and evaluate the pervasiveness and impact of biases across studies.

To address these gaps, this study aims to conduct a systematic review that synthesizes and evaluates existing machine-learning models for detecting depression on social media. The specific objectives are:
\begin{enumerate}
\item Examine the effectiveness of machine learning and deep learning models by focusing on bias present in sampling, data preprocessing, model construction, fine-tuning, evaluation, and comparison, as well as the challenges associated with model generalizability across different social media platforms.
\item Explore methodological challenges, including those unique to mental health detection—such as handling class imbalance where depressive posts are the minority and preprocessing for sentiment analysis involving negations. Additionally, more general machine learning challenges, like improving model evaluation techniques and addressing data biases related to language and platform-specific factors, also persist. It is important to recognize that most of these biases are unintentional, either from practical challenges or from a lack of standardized guidelines for applying machine learning to mental health detection. By addressing these biases, the review aims to provide insights and strategies to mitigate these unintended biases, advancing the development of more reliable and generalizable models.
\item Provide recommendations for future research to enhance the reliability and applicability of machine learning models in mental health detection. These insights aim to inform strategies that improve early intervention efforts and contribute to the development of more robust, generalizable, and ethically sound machine learning applications. In doing so, the review seeks to provide guidance that fills the gap left by current practice, where a lack of formal guidelines has sometimes led to the persistence of unintended biases.
\end{enumerate}

By addressing these objectives, this review seeks to provide a comprehensive understanding of the current practices and limitations within the field. The findings aim to guide future research and development into more robust, generalizable, and ethical applications of machine-learning models for mental health detection using social media data. In the following sections, we will first examine the methodologies and models used across studies, followed by an analysis of common biases and limitations. We will conclude with a discussion on best practices and recommendations for advancing the field.

\section{Methodology}

\subsection{Search Strategy}
The search focused on publications on machine learning and deep learning models for detecting depression and other mental health conditions using social media data, primarily from platforms like Twitter, Facebook, and Reddit. To identify relevant studies, a systematic search was conducted across multiple academic databases including PubMed, ACM, and IEEE Xplore, with Google Scholar used for additional sources. The search included combinations of `machine learning,' `deep learning,' `artificial intelligence,' `social media,' `Twitter,' `Facebook,' `Reddit,' `depression,' `sentiment analysis,' and `mental health.' To broaden the scope of the search, additional terms such as `anxiety,' `mental disorders,' `neural networks,' and `supervised learning' were included. The search process was carried out from June to July 2024.

The search strategy was structured around three main categories: social media platforms (e.g., `social media,' `Twitter,' `Facebook,' `Reddit'), mental health topics (e.g., `depression,' `sentiment analysis'), and machine learning and data analysis techniques (e.g., `machine learning,' `deep learning,' `artificial intelligence'). The comprehensive search query\footnote{The search query used the term `Twitter' to align with the naming convention at the time of the review, which covered literature up to June/July 2024. Twitter was rebranded as `X' after this period. The search algorithm was adjusted to include both `Twitter' and `X' where applicable to ensure coverage of relevant results under the new name. However, no additional papers published up to June/July 2024 were identified using the term `X.' Notably, one manuscript, \citeA{jamali2023momentary}, included both terms.} formulated for this review is:

\begin{verbatim}
((social media OR 'Twitter' OR 'Facebook' OR 'Reddit') AND ('depression' OR 
'sentiment analysis' OR 'mental health' OR 'anxiety' OR 'mental disorders') 
AND ('machine learning' OR 'deep learning' OR 'artificial intelligence' OR 
'neural networks' OR 'supervised learning')).
\end{verbatim}

\subsection{Inclusion and Exclusion Criteria}
To be included in this review, studies needed to meet the following criteria:
\begin{itemize}
\item \textbf{Publication Date:} Studies published after 2010 were included to ensure contemporary research and methods were considered.
\item \textbf{Language:} Only studies published in English were included.
\item \textbf{Research Focus:} The study must use machine learning or deep learning models for detecting depression or other mental health conditions, with a particular focus on analyzing data from social media platforms like Twitter, Facebook, or Reddit.
\item \textbf{Study Type:} The review included primary research articles, specifically those that involved data-driven analyses.
\end{itemize}

Studies were excluded based on the following criteria:
\begin{itemize}
\item \textbf{Publication Type:} Review articles, systematic reviews, conference abstracts, editorials, opinion pieces, and non-peer-reviewed literature were excluded.
\item \textbf{Scope:} Studies not directly focused on mental health detection through social media or not employing machine learning models were excluded.
\item \textbf{Methodology:} Studies that did not directly employ machine learning or deep learning and applied solely on quantitative analysis were excluded.
\end{itemize}

\subsection{Study Selection Process}
The selection process was conducted in three stages to ensure a rigorous and unbiased review of relevant studies. The process, which followed the search process that concluded in July 2024, lasted until August 2024. 
\begin{enumerate}
\item \textbf{Initial Identification:} Duplicates were removed, and an initial screening was conducted based on titles and abstracts to filter out irrelevant studies. All authors contributed to this step.
\item \textbf{Title and Abstract Screening:} An independent review was conducted by two authors, Y.T. and J.D., to assess the relevance of studies based on their titles and abstracts. Both authors have expertise in machine learning and mental health research, ensuring a thorough evaluation. Any discrepancies in their assessments were discussed and resolved to ensure a consistent screening process.
\item \textbf{Full-Text Screening:} A comprehensive review of the full texts of selected studies was conducted. Any disagreements were resolved through discussion to maintain an unbiased selection process. Additionally, relevant studies identified through references in full-text articles were included for consideration. All authors contributed to this step.
\end{enumerate}

\subsection{Data Extraction and Analysis}
The data extraction process involved using a standardized form to systematically capture detailed information from each selected study. The form included fields to record author names, study titles, publication journals, and publication years. It also documented the study designs, settings, and sample sizes, alongside specific inclusion and exclusion criteria. In addition, the form provided details on the machine learning models employed, the social media platforms analyzed (such as Twitter, Facebook, and Weibo), and the primary and secondary outcomes measured. Additionally, performance metrics, including accuracy, precision, recall, F1 score, and Area Under the Receiver Operating Characteristic (AUROC)\footnote{Accuracy measures the proportion of correctly classified instances among all instances. Precision focuses on the correctness of positive predictions, while recall measures the ability to identify actual positive cases. Both F1-score and Area Under the Receiver Operating Characteristic Curve (AUROC) are composite metrics that combine aspects of precision and recall to evaluate the performance of models. A detailed explanation of these metrics is provided in Section~\ref{sec:metric}}, which were collected when applicable.

Special attention was given to identifying potential sources of bias, study limitations, and funding sources, ensuring a comprehensive overview of each study's context and reliability. Table~\ref{tab1:data_extraction} below outlines the key categories and details included in the data extraction form.

\begin{table}[ht]
\centering
\caption{Key Data Extraction Categories for Systematic Review.}
\resizebox{\textwidth}{!}{%
\begin{tabular}{l|p{10cm}} 
\hline
\textbf{Category} & \textbf{Details} \\
\hline
Study Details & Title, Authors, Year of Publication, Journal or Source, DOI or URL \\
\hline
Research Objectives & Purpose of the Study, Research Questions or Hypotheses \\
\hline
Methodological Aspects & Study Design, Settings, Sample Sizes, Inclusion and Exclusion Criteria, Data Collection Methods, ML/DL Models Employed \\
\hline
Criteria Applied & Data included, e.g., publicly available tweets, specific language posts. Data excluded, e.g., private or insufficiently detailed posts \\
\hline
Performance Metrics & Metrics Used (e.g., Accuracy, Precision, Recall, F1-score, AUROC, etc.) \\
\hline
Bias Evaluation & Data Collection and Preprocessing, Model Development and Tuning, Model Evaluation and Reporting \\
\hline
Additional Information & Confounding Factors, Study Limitations, Ethical Considerations, Funding Sources \\
\hline
\end{tabular}%
}
\label{tab1:data_extraction}
\end{table}

\subsection{Analytical Methods Used to Synthesize Findings}
The extracted data were synthesized using a narrative approach, systematically examining each aspect of the machine learning lifecycle—sampling, data preprocessing, model construction, tuning, evaluation, comparison, and reporting—across the selected studies. This synthesis involved reviewing how studies approached sampling and data preprocessing, examining their approaches to model construction and tuning, and assessing model evaluation and comparison based on quantitative metrics such as accuracy, precision, recall, F1 scores, and AUROCs. For each stage, we summarized the methodologies employed by the studies and identified potential biases with established tools. This comprehensive approach provided insights into the current state of research, highlighting areas for future investigation to enhance the accuracy, generalizability, and applicability of machine learning models in this field.

\subsection{Systematic Review Registration}
This systematic review has been registered in the International Prospective Register of Systematic Reviews (PROSPERO) database under the title \textit{Systematic Review of Machine Learning and Deep Learning Algorithms for Detecting Depression and Mental Health Conditions on Social Media} (ID: 617763). The registration has been approved.

\section{Results}
\subsection{Study Selection}
The search process began by identifying a total of 328 studies from three databases: 192 from Google Scholar, 101 from PubMed, and 35 from IEEE Xplore. After removing 57 duplicate studies, 271 unique titles and abstracts were retained for screening. During the title and abstract review, 174 studies were excluded. These exclusions were due to issues related to methodology (53 studies), scope (77 studies), and publication type (44 studies). This left 97 full-text studies to be reviewed in detail.

Upon reviewing the full texts, another 50 publications were excluded. The reasons for exclusion included being outside the scope or irrelevant (32 studies), methodological concerns (6 studies), publication type (9 studies), and unavailability (3 studies). Ultimately, 47 studies were included in the final narrative synthesis.

\begin{figure}[!htb]
\centering
\includegraphics[width=0.9\linewidth]{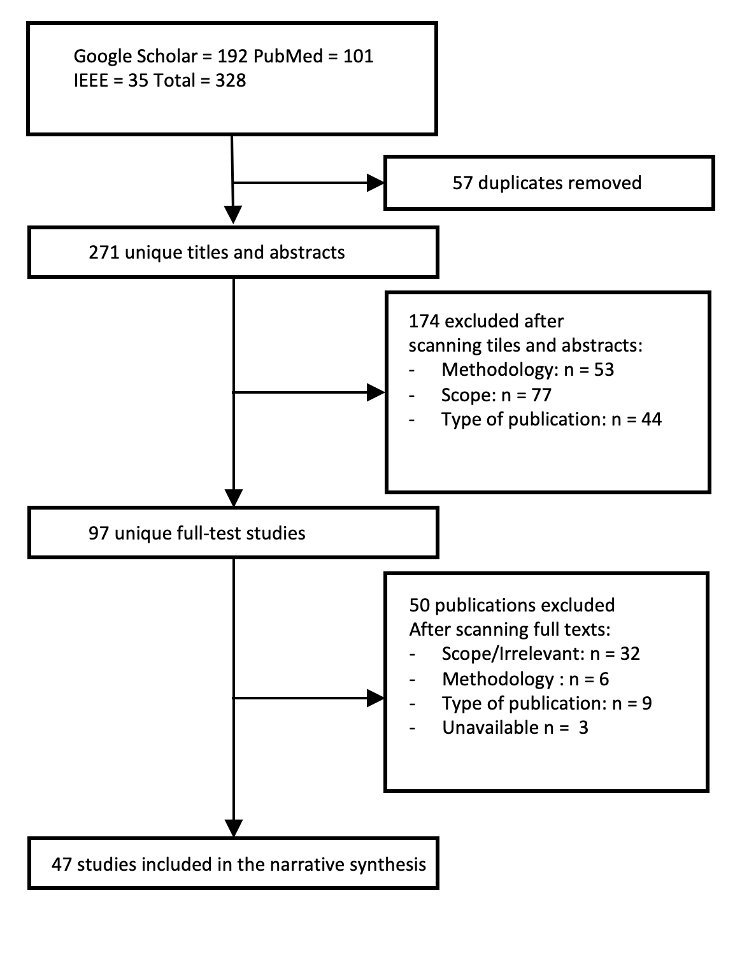}
\caption{PRISMA Flow Diagram of Study Selection Process for Systematic Review on Machine Learning Models for Depression Detection Using Social Media Data}
\label{fig:PRISMA}
\end{figure}

Figure~\ref{fig:PRISMA} outlines how the initial pool of studies was refined down to the most relevant research for inclusion.

\subsection{Characteristics of Included Studies}
In this systematic review, key details of all 47 included studies, as summarized in Table~\ref{tab1:data_extraction}, are provided in an online supplementary document. The majority of studies focused on Twitter (32 studies), Reddit (8 studies), and Facebook (7 studies). Additionally, one study examined Blued, a platform for MSM communities, and another focused on Indian social networking sites (SNS). Notably, 8 studies (17.02\%) analyzed data from multiple platforms. Upon further examination, the datasets used in these 47 studies were found to be independent. The most commonly used models included traditional machine learning approaches such as Support Vector Machines (SVM) (19 studies), tree-based models (e.g., Decision Trees in 6 studies, Random Forests in 13 studies, and eXtreme Gradient Boosting (XGBoost) in 3 studies), and Logistic Regression (6 studies). Some studies also utilized deep learning models, including Convolutional Neural Networks (CNNs) (9 studies), Long Short-Term Memory (LSTM) networks (5 studies), and Bidirectional Encoder Representations from Transformers (BERT) (9 studies) for depression detection.

\subsection{Methodological Quality and Risk of Bias}
The risk of bias in the studies included in this systematic review was assessed using the Prediction model Risk Of Bias Assessment Tool (PROBAST) \cite{wolff2019probast}. PROBAST is a structured tool designed to assess the risk of bias and applicability of prediction models. It evaluates four key domains: participants, predictors, outcomes, and analysis, ensuring methodological rigor in studies. This tool provides a systematic framework for identifying biases and limitations in prediction models, offering critical insights into their validity and applicability. PROBAST is particularly relevant in this systematic review as it allows for a comprehensive assessment of potential methodological biases throughout the machine learning lifecycle, including data collection, preprocessing, model development, and evaluation. By identifying biases in these areas, the tool supports a rigorous evaluation of the reliability and generalizability of machine learning models used for mental health detection on social media. In addition to its role in assessing the risk of bias, PROBAST was used to evaluate transparency and completeness in the reporting of study methodologies and findings. The assessment covered 20 structured questions across the four domains, as detailed in Table~\ref{tab2:bias_evaluation}. By incorporating PROBAST, this review identifies methodological weaknesses in the included studies, assesses their implications for the validity of findings, and evaluates the overall applicability of machine learning models used for mental health detection on social media. This ensures a thorough understanding of bias and enhances the reliability of the review’s conclusions.

\begin{table}[ht]
\centering
\caption{Bias Evaluation Questions for Each Domain.}
\resizebox{\textwidth}{!}{%
\begin{tabular}{l|p{10cm}} 
\hline
\textbf{Domain} & \textbf{Evaluation Questions} \\
\hline
Sample Selection and Representativeness & 
Q1. What is the sample used in this study, including the platform, sampling criteria, and sampling method? \newline
Q2. Does the sample represent the target population of social media users or posts? \\
\hline
Data Preprocessing & 
Q3. Did the study specify its approach to handling negative words when using traditional or machine learning methods for sentiment analysis? \\
\hline
Model Development & 
Q4. Did this study report hyperparameters? \newline
Q5. If reported, did this study tune (optimize) hyperparameters or use default settings? \newline
Q6. If tuned hyperparameters in this study, was this done on all models mentioned in the study? \\
\hline
Model Evaluation & 
Q7. Did the study divide the dataset into training, validation, and test sets, and were the reported metrics based only on training data? \newline
Q8. What evaluation metric was used in this study? \newline
Q9. Is the evaluation metric appropriate for this context (i.e., class-imbalanced settings)? \newline
Q10. If the study used accuracy as an evaluation metric, did it mention preprocessing steps to address class imbalance? \\
\hline
\end{tabular}%
}\label{tab2:bias_evaluation}
\end{table}

\subsection{Sample Selection and Representativeness (Q1 \& Q2)}
The reviewed studies employed diverse sampling methods across various social media platforms, primarily focusing on Twitter (63.8\%) with additional data from Reddit (23.5\%), Facebook (8.5\%), and other social media (2.1\%). Most studies (around 80\%) used non-probability sampling techniques, such as convenience sampling or keyword filtering, often utilizing APIs (e.g., Twitter API, Reddit API) to filter posts by specific mental health-related keywords like `depression' or `\#MentalHealth,' or leveraging pre-existing datasets from repositories like Kaggle.

The diversity in sampling criteria, sample sizes, demographic details, language focus, and geographic regions across the studies introduces potential biases. Sample sizes and levels of representation varied significantly among the studies, from small-scale studies (e.g., Study \#46, which analyzed 4,124 Facebook posts from 43 undergraduate students with pre-specified criteria from the U.S.) to large-scale analyses (e.g., Study \#5, which analyzed 56,411,200 tweets from 70,000 users across seven major U.S. cities). Many studies lacked detailed demographic information. The majority of studies focused predominantly on English-language posts, which are commonly associated with specific regions such as the U.S., U.K., Japan, Spain, and Portugal (although geographic information was explicitly reported in only about one-third of the studies) limiting the generalizability of the findings. Only a few studies examined posts in other languages, like Study \#15, which analyzed Arabic tweets. Even within these regions and language-specific studies, demographic distribution was not always fully balanced. For example, Study \#1 reported a mean participant age of 30.5 years (ranging from 18 to 68) and had a slight overrepresentation of female participants at 66.4\%.

The non-representative sampling approaches observed across studies suggest limited generalizability to broader social media user populations. The primary biases identified include:

\begin{itemize}
\item \textbf{Platform Bias:} The predominance of Twitter (63.8\%) over other platforms means that findings may not represent behaviors on platforms like Facebook, Instagram, or Reddit. As suggested by \citeA{olteanu2019social}, utilizing multi-platform data can reduce platform-specific biases and provide a more comprehensive view of user behaviors. However, while multi-platform data broaden the scope and reduce single-platform bias, platform-specific user demographics and engagement patterns may affect generalizability, with some platforms carrying more weight due to larger user bases or data volume.
\item \textbf{Selection Bias:} Some studies relied on keyword-based sampling, which may overlook users not explicitly mentioning mental health. Study \#7, for instance, searched for tweets containing `I was diagnosed with depression.' As suggested by \citeA{morstatter2013sample}, combining keyword-based and random sampling can capture a broader range of user behaviors and discussions. Additionally, the limitations of Twitter’s API exacerbate platform-specific challenges. As highlighted by \citeA{morstatter2013sample}, Twitter’s API does not provide access to all user-generated content, raising concerns about whether sampled data is representative of the platform’s overall activity. This issue may lead to incomplete or skewed representations of user behavior, particularly in studies relying solely on API data. Researchers must critically evaluate the validity of conclusions drawn from API-retrieved data and consider combining multiple sampling strategies to mitigate such biases.
\item \textbf{Language Bias:} The overwhelming focus on English-language content (over 90\%) excludes insights from non-English-speaking communities, limiting the generalizability of findings across diverse linguistic groups. For instance, Study \#15 was one of the few that analyzed non-English tweets, indicating the rarity of multilingual studies in this field. To address this, \citeA{danet2007multilingual} recommended leveraging multilingual analysis methods, such as machine translation, or employing multilingual research teams to capture a more diverse linguistic landscape.
\item \textbf{Geographic Bias:} While explicit geographic information was reported in only about one-third of the studies, the predominance of English-language posts suggests an implicit bias toward regions where English is the primary language, such as the U.S., U.K., and other English-speaking countries. Among the studies that reported geographic information, this predominance was evident. For example, Study \#5 analyzed tweets from seven major U.S. cities, and Study \#19 focused on Twitter users in Spain and Portugal. \citeA{hargittai2015bigdata} suggested broadening the geographic scope to better represent global populations and avoid region-specific findings.
\item \textbf{Self-selection Bias:} Platforms like Mechanical Turk (MTurk) or Clickworker, used in some studies (e.g., Studies \#45 and \#1, respectively), may attract specific demographic or employment profiles (e.g., higher digital literacy, particular age ranges, or specific socioeconomic statuses), affecting generalizability. While \citeA{chandler2016crowdsourced} assessed the use of MTurk as a crowdsourcing tool, highlighting limitations in participant diversity and representativeness, which may skew results and underscore the need for multiple recruitment sources and stratified sampling for better generalizability.
\end{itemize}

In summary, no study in the review provided a fully representative sample of all social media users or posts. Key limitations include platform-specific focus (mostly Twitter), heavy reliance on non-probability sampling techniques (e.g., approximately 80\% of the studies utilized convenience sampling or keyword filtering), and geographic and linguistic constraints. Notably, over 90\% of the studies themselves acknowledged these limitations, recognizing the challenges of achieving representativeness in social media research. These limitations are, to a large extent, unavoidable due to the nature of social media platforms and the constraints of current data collection methodologies. This underscores the need for ongoing efforts to develop more sophisticated sampling techniques and analytical methods to mitigate these biases.

Similarly, some studies explicitly stated that their findings were intended to represent only specific populations. For instance, Study \#8 and Study \#21 focused on users discussing mental health or particular demographic groups on specific platforms. These limitations significantly impact the generalizability of findings to the broader population of social media users. Future research should strive for more diverse and representative sampling across platforms, languages, and geographic regions to enhance the applicability of results in the field of mental health and social media research.

\subsection{Data Preprocessing with Focus on Negative Words Handling (Q3)}
Across all studies, several common preprocessing tasks were consistently performed. Tokenization was conducted in all studies to break text into individual words or tokens, and text normalization steps included converting text to lowercase, as well as removing punctuation, URLs, and special characters. Many studies also performed stop-word removal to eliminate common words that are generally not informative for modeling. Additionally, some studies applied stemming and lemmatization to reduce words to their base or root forms, thereby unifying different morphological variants. Feature extraction techniques such as Bag of Words (BoW)\footnote{BoW represents text as a vector by creating a vocabulary of all unique words in a corpus and counting the frequency of each word in a document. While simple and effective, BoW disregards word order and context, treating documents as collections of independent words.} \cite{harris1954bow}, Term Frequency-Inverse Document Frequency (TF-IDF)\footnote{TF-IDF evaluates the importance of a word in a document relative to a collection of documents. It combines term frequency (how often a word appears in a document) with inverse document frequency (reducing the weight of common words that appear across many documents). This technique highlights terms that are more informative for classification or clustering tasks.} \cite{salton1988tfidf}, and various word embedding methods were widely used to represent textual data numerically for modeling purposes.

While these standard preprocessing steps were broadly applied, certain aspects of sentiment analysis in mental health detection require additional attention. One such aspect is the effective handling of negative words, which is crucial for accurately interpreting sentiment and emotional tone, especially within this context. Among the 47 reviewed studies, approaches to negative words varied significantly:

First, only a minority of studies (11 out of 47 studies, approximately 23\%) explicitly addressed negative words or negations in their preprocessing steps. Methods included standardizing all negative words to a basic form, like `not,'' during preprocessing, which simplifies the representation of negations and improves sentiment recognition (e.g., Studies \#3 and \#34). Some studies quantified negative words as features by calculating metrics such as the user-specific average number of negative words per post. This metric captures the frequency of negative expressions per user and is then used as input for machine learning models to identify depressive emotions (e.g., Study \#21). Others (e.g., Study \#25) assigned a weight of $-1$ to negative adverbs to account for their inversion effect on sentence sentiment, ensuring more accurate sentiment quantification. Moreover, several studies employed specific methods for managing negations within their sentiment analysis frameworks. For example, some studies used sentiment analysis tools like TextBlob to determine the polarity of words in context, identifying negative words as indicators of depressive symptoms (e.g., Study \#31). Others incorporated linguistic inquiry and word count (LIWC) categories related to negations and negative emotions, indirectly addressing negations through predefined lexicon categories (Studies \#1, \#40, \#42, \#46, and \#47).

The importance of negation handling has also been recognized in studies currently under review. For instance, Study \#6 specifically explored the role of negation preprocessing in sentiment analysis for depression detection. By comparing datasets with and without negation handling, the authors demonstrated that addressing negations can significantly improve the accuracy of both sentiment analysis and depression detection, underscoring the need to address them in preprocessing. This study highlights the critical need for comprehensive negation handling in preprocessing to enhance the reliability of machine learning models in mental health contexts.

Second, a subset of studies (9 out of 47 studies, approximately 19\%) did not explicitly handle negative words but employed advanced language models capable of inherently managing negations due to their contextual understanding, such as transformer-based models like Bidirectional Encoder Representations from Transformers (BERT) \cite{devlin2018bert} and Mental Health BERT (MentalBERT) \cite{ji2022mentalbert} (e.g., Studies \#8, \#9, \#15, \#16, and \#39). These transformer-based models can capture the context of negations by processing text bi-directionally without explicit preprocessing steps. Other studies used attention mechanisms\footnote{Attention mechanisms allow models to focus on specific parts of the input data by assigning different weights to different elements. This enables the model to capture and utilize relevant contextual information more effectively during processing.} \cite{vaswani2017attention} with word embeddings, such as attention layers combined with Global Vectors for Word Representation (GloVe) embeddings \cite{pennington2014glove}, allowing models to inherently understand and assign appropriate weights to negations through contextual embeddings (e.g., Studies \#7, \#10, and \#13). Additionally, Embeddings from Language Models (ELMo) \cite{peters2018elmo}, which capture the entire context of a word within a sentence, was also noted as a method that could capture the effect of negative words without explicit handling (Study \#45).

However, the majority (27 out of 47 studies, approximately 57\%) neither explicitly addressed negative words in their preprocessing nor used models inherently capable of handling negations (i.e., Studies \#2, \#4, \#5, \#11, \#12, \#14, \#17, \#18, \#19, \#20, \#22, \#23, \#24, \#26, \#27, \#28, \#29, \#30, \#32, \#33, \#35, \#36, \#37, \#38, \#41, \#43, and \#44). These studies primarily focused on standard preprocessing tasks (e.g., tokenization, lowercasing, stop-word removal, stemming, and lemmatization), feature extraction methods (e.g., TF-IDF, BoW), and basic word embeddings (e.g., Word to Vector [Word2Vec]), without any special consideration for negations.

The impact on model performance and potential bias varied depending on how negative words were handled. Studies that explicitly addressed negative word handling reported improvements in model accuracy and a more nuanced understanding of sentiment \cite{helmy2024depression}. Proper handling of negations allowed these models to correctly interpret phrases where negations invert the sentiment (e.g., `not happy'' versus `happy''), leading to more reliable results. In contrast, studies that did not explicitly account for negative words risked misinterpreting negated expressions, introducing bias into their findings. This oversight can cause models to incorrectly assign positive sentiment to negated negative expressions or vice versa, thus skewing the analysis. Such biases can significantly affect the overall performance and generalizability of the models, particularly in sensitive applications like depression detection. While some studies used advanced models capable of inherently handling negations (e.g., Studies \#7, \#8, \#9, \#10, \#13, \#15, \#16, \#39, and \#45), reliance solely on the model's ability without explicit preprocessing might not capture all nuances of negations. Explicitly addressing negations can further enhance model performance, even when using sophisticated language models \cite{khandeparkar2020negbert}. Therefore, integrating both advanced modeling techniques and careful preprocessing of negative words may provide the most effective approach.

In summary, the review highlights a significant gap in the explicit handling of negative words in data preprocessing among studies focused on sentiment analysis and related fields. Proper management of negations is crucial, as it can substantially impact both model accuracy and reliability. Without adequately handling negative words, models may introduce bias and reduce their effectiveness, particularly in applications such as mental analysis and depression detection, where understanding sentiment nuances is critical. Future studies should prioritize the inclusion of explicit negation handling techniques within their preprocessing pipelines to enhance model performance and ensure more accurate interpretations of textual data.

\subsection{Model Development}
\subsubsection{Hyperparameter Tuning (Q3, Q4 \& Q5)}
Hyperparameters are external configurations set before the training process of machine learning models. Unlike model parameters, which are learned from the data during training, hyperparameters govern the learning process itself, such as the learning rate, regularization strength, and the number of hidden layers. Proper hyperparameter tuning ensures optimal model performance by balancing underfitting and overfitting, thus improving the model's ability to generalize to unseen data. Hyperparameter tuning is a critical aspect of optimizing machine learning models, directly impacting their performance and reliability. Our evaluation of the 47 reviewed studies focused on whether the studies reported their hyperparameters, the extent to which these hyperparameters were optimized, and whether tuning was applied consistently across all models within each study.

In particular, 27 studies (approximately 60\%) reported using hyperparameters, but not all of them performed proper tuning. Only a limited number of studies ensured consistent tuning across all models, with many opting for default settings or tuning only specific models, leaving significant performance potential unexplored \cite{yang2020optimization}. This practice suggests that while hyperparameters are acknowledged by researchers, there is still a notable gap in their comprehensive and consistent optimization across studies. The breakdown of hyperparameter reporting and tuning practices is presented in Table~\ref{tab3:hyperparameter_reporting}.

\begin{table}[ht]
\centering
\caption{Hyperparameter Reporting and Tuning Practices in Reviewed Studies.}
\resizebox{\textwidth}{!}{%
\begin{tabular}{l|c|p{7cm}} % Use fixed width for the third column
\hline
\textbf{Hyperparameter Reporting} & \textbf{Number (\%) of Studies} & \textbf{Studies \#} \\
\hline
Reported \& Tuned for All Models & 13 (27.7\%) & \#11, \#12, \#13, \#16, \#18, 
\#21, \#22, \#25, \#26, \#28, \#33, \#45, \#47 \\
\hline
Reported but Partially Tuned & 4 (8.5\%) & \#1, \#8, \#15, \#23 \\
\hline
Reported but Not Tuned & 11 (23.4\%) & \#3, \#4, \#7, \#9, \#10, \#31, 
\#36, \#39, \#40, \#41, \#43 \\
\hline
Not Reported or Tuned & 19 (40.4\%) & \#2, \#5, \#6, \#14, \#17, 
\#19, \#20, \#24, \#27, \#29, 
\#30, \#32, \#34, \#35, \#37, 
\#38, \#42, \#44, \#46 \\
\hline
\end{tabular}%
}
\label{tab3:hyperparameter_reporting}
\end{table}

The absence of consistent hyperparameter tuning can result in suboptimal model performance, reduced generalizability, or biased model comparisons. Key hyperparameters such as learning rate, regularization terms, or the number of hidden layers directly impact a model’s training process and final accuracy \cite{probst2019hyperparameters, Mantovani2015}. Without proper tuning, models may overfit, meaning they perform well on training data but poorly on unseen data, or underfit, failing to capture the complexity of the data altogether. For example, Study \#2 did not report any tuning, which likely affected its model's ability to generalize to unseen data, leading to reduced model performance.

When only some models are tuned, comparisons across models become biased, as those with optimized hyperparameters gain an undue advantage. In Study \#1, for instance, the Elastic Net model had its hyperparameters tuned, while other models, such as random forest, were left with default settings. This discrepancy can misleadingly suggest the superiority of the Elastic Net model due to tuning alone, rather than any inherent advantage in its architecture, leading to biased model comparisons.

A significant proportion of studies did not report hyperparameter tuning (approximately 40\%) or failed to consistently tune them across all models (approximately 32\%), which compromises the validity of their findings. For example, Studies \#2 and \#4 used default settings and missed opportunities to enhance performance, while Study \#1 tuned hyperparameters for only one model, resulting in biased comparisons. Proper hyperparameter tuning is essential to avoid issues like overfitting or underfitting. Consistent tuning across all models ensures fair comparisons and enhances result validity.

Providing detailed descriptions of hyperparameter settings and optimization processes enhances transparency and reproducibility. Standardized tuning protocols, such as grid search, random search, or Bayesian optimization, should be employed to explore optimal configurations. Clearly documenting tuning strategies and any challenges encountered will provide valuable context for interpreting model performance results and strengthen the credibility of future machine learning studies. Future research should prioritize consistent tuning strategies and detailed reporting to enhance the credibility and reproducibility of their machine learning studies.

\subsubsection{Data Partitioning (Q6)}
Proper data partitioning is fundamental to developing robust machine learning models that generalize well to unseen data. Typically, datasets are divided into three subsets: the training set, used to train the model and learn patterns; the validation set, used to fine-tune hyperparameters and avoid overfitting; and the test set, reserved for evaluating the model's final performance on unseen data. Of the 47 reviewed studies, 32 studies (approximately 68\%) adhered to recommended machine learning protocols by appropriately dividing their datasets into training, validation, and test sets or by employing cross-validation techniques. The breakdown of data partitioning practices is summarized in Table~\ref{tab4:data_partitioning}.

Among the studies that explicitly partitioned their datasets, such as Studies \#1, \#6, and \#7, performance metrics were reported based on the test sets, adhering to the best practices outlined by \citeA{goodfellow2016deeplearning}. By evaluating their models on unseen data, they ensured that the models' performance accurately reflected their generalizability.

\begin{table}[ht]
\centering
\caption{Summary of Data Partitioning Practices Across Reviewed Studies.}
\resizebox{\textwidth}{!}{%
\begin{tabular}{l|c|p{7cm}} % Adjust the width (e.g., 12cm) of the third column
\hline
\textbf{Data Partitioning Practices} & \textbf{Number (\%) of Studies} & \textbf{Studies \#} \\
\hline
Training/Validation/Test Split & 32 (68.1\%) & \#1, \#6, \#7, \#8, \#10, \#11, \#13, \#15, \#16, \#17, \#18, \#19, \#21, \#22, \#23, \#25, \#26, \#28, \#29, \#30, \#32, \#33, \#34, \#35, \#36, \#40, \#41, \#42, \#43, \#45, \#46, \#47 \\
\hline
Cross-validation without Traditional Split & 7 (14.9\%) & \#3, \#4, \#14, \#24, \#38, \#39, \#44 \\
\hline
Inadequate or Unreported Partitioning & 8 (17.0\%) & \#2, \#5, \#9, \#12, \#20, \#27, \#31, \#37 \\
\hline
\end{tabular}%
}
\label{tab4:data_partitioning}
\end{table}

Seven studies used cross-validation methods instead of a traditional train/validation/test split. Techniques like k-fold cross-validation provide a robust assessment of a model's ability to generalize by iteratively training and testing on different subsets of the dataset \cite{hastie2009elements}. For instance, Study \#39 utilized 5-fold cross-validation, where the dataset was divided into five subsets, with each subset used as a test set once while the remaining subsets formed the training set. The reported metrics—Positive Predictive Value (PPV), Sensitivity, and F1 Score—were averaged across the five test folds in the cross-validation process, ensuring that evaluation was based on separate test data rather than solely on the training data.

Conversely, as shown in Table~\ref{tab4:data_partitioning}, approximately 17\% of studies (8 out of 47) did not report sufficient details on data partitioning or did not employ partitioning techniques. For example, Study \#2 provided limited information about its dataset division and did not elaborate on how model performance was evaluated, while Study \#5 applied pre-existing models without conducting new data partitioning or validation within their analysis, thereby limiting the validity of their performance assessments.

Inadequate data partitioning practices introduce significant risks of bias, particularly overfitting. Models that lack proper data division tend to memorize the training data, leading to overly optimistic performance metrics that do not accurately reflect real-world applicability \cite{bishop2006ml}.

According to \citeA{ng2018ml}, proper validation and testing sets are crucial for assessing generalization and preventing overfitting. Without these, models may appear overly effective due to inflated performance metrics, misleading when applied beyond the training context. For example, studies that evaluated models solely on training data, such as Studies \#2 and \#5, likely overestimate their real-world performance.

In summary, while the majority of the reviewed studies adhered to best practices in data partitioning—thereby enhancing the credibility and generalizability of their findings—a significant minority did not. The lack of proper data partitioning in approximately 17\% of studies introduces risks of bias, underscoring the need for more rigorous practices. For the development of robust models, future research should consistently apply proper data partitioning and report performance based on validation or test sets to provide accurate, unbiased evaluations. Transparent data partitioning and evaluation reporting, as emphasized by \citeA{bishop2006ml} and \citeA{goodfellow2016deeplearning}, is fundamental to enhancing reproducibility and reliability in machine learning research. By incorporating these practices, researchers can enhance the reliability of their models, ensure that findings are both valid and applicable in real-world scenarios, and contribute to the advancement of the field.

\subsection{Model Evaluation: Evaluation Metrics for Imbalanced Class Scenarios (Q8, Q9 \& Q10)}\label{sec:metric}
In the domain of depression-related emotion detection, datasets often exhibit significant class imbalance, with non-depressed cases vastly outnumbering depressed ones. This imbalance poses challenges for model evaluation, as traditional metrics like accuracy can be misleading. According to \citeA{he2009imbalanced}, accuracy may not adequately reflect a model's performance in imbalanced scenarios because a model could achieve high accuracy by simply predicting the majority class. Therefore, metrics such as recall, precision, F1 score, and Area Under the Receiver Operating Characteristic Curve (AUROC or AUC) are preferred, as they provide a more balanced evaluation by accounting for both false positives and false negatives. \citeA{he2013assessment, japkowicz2002imbalance} further emphasize the necessity of using these metrics, arguing that they are crucial for a comprehensive assessment of model performance in the presence of class imbalance.

In the context of depression detection, recall, measures the proportion of actual positive cases (individuals with depression) that the model correctly identifies (i.e., $\text{Recall} = \frac{\text{True Positive}}{\text{True Positive} + \text{False Negative}}
$), is particularly important. A high recall indicates that the model is successfully identifying most individuals who are truly depressed (true positives), although this often comes at the cost of more false positives, where individuals without depression are incorrectly flagged as depressed. Failing to identify someone who is depressed (a false negative) could have serious consequences, as it may result in a missed opportunity to provide help or intervention. Therefore, prioritizing recall ensures that the model captures as many true positive cases as possible, even if it risks increasing false positives. In this context, minimizing false negatives is often a higher priority, given the potential implications for those who might otherwise go undiagnosed and unsupported \cite{Bradford2024Diagnostic}.

Precision, on the other hand, measures the proportion of positive predictions that are correct (i.e., $\text{Precision} = \frac{\text{True Positive}}{\text{True Positive} + \text{False Positive}}$), highlighting the model’s ability to avoid false positives. In depression detection, a low precision score indicates a high rate of false positives, where individuals who are not depressed are incorrectly labeled as depressed. This could lead to unnecessary concern or even stigmatization for those wrongly flagged. While high precision is desirable to avoid false alarms, an overly strict focus on precision could inadvertently lower recall, leading to more false negatives. Therefore, balancing precision and recall is essential to ensure that the model is not only identifying true cases of depression but also minimizing the number of false alarms. This balance is particularly critical in applications where both false negatives (missing a depressed individual) and false positives (incorrectly flagging someone as depressed) carry significant consequences \cite{Bradford2024Diagnostic}.

The F1 score, representing the harmonic mean of precision and recall, provides a balanced measure of both recall and precision. It is particularly useful in imbalanced datasets, where a balance between recall and precision is essential.

Finally, AUROC measures the model's ability to distinguish between positive and negative classes across different threshold settings, providing a comprehensive view of the model's discriminatory power. A higher AUROC indicates a better capability of distinguishing between depressed and non-depressed individuals, making it a robust metric for evaluating models in this domain. Among the 47 studies reviewed, approximately 35 (Studies \#1, \#3, \#6, \#7, \#8, \#13, \#14, \#15, \#16, \#17, \#19, \#21, \#22, \#23, \#25, \#26, \#27, \#28, \#29, \#30, \#31, \#32, \#33, \#34, \#35, \#36, \#37, \#39, \#40, \#41, \#42, \#43, \#44, \#45, \#46) utilized these preferred metrics. For example, Study \#6, `Depression Detection for Twitter Users Using Sentiment Analysis in English and Arabic Tweets,' employed precision, recall, F1 score, and AUC to evaluate their models, acknowledging the importance of these metrics for imbalanced data. Similarly, Study \#42, `Classification of Helpful Comments on Online Suicide Watch Forums,' emphasized recall as a key metric in evaluating their model's effectiveness in identifying individuals at risk.

Other than the utilization of preferred metrics, an alternative way to address imbalanced data involves implementing data balancing techniques, including resampling and reweighting. For instance, Study \#6 employed dynamic sampling methods, such as oversampling the minority class and undersampling the majority class, to balance the dataset. This approach ensured that the model had sufficient exposure to both classes before model construction and evaluation. Similarly, Study \#41, `A Deep Learning Model for Detecting Mental Illness from User Content on Social Media,' used Synthetic Minority Oversampling Technique (SMOTE) to enhance the representation of the minority class, leading to improved classification performance, particularly for underrepresented classes.

Notably, some studies (Studies \#3, \#6, \#13, \#15, \#34, \#40, \#41, \#42, \#43) applied both data balancing techniques and preferred evaluation metrics together to comprehensively address the class imbalance. For example, `Explainable Depression Detection with Multi-Aspect Features Using a Hybrid Deep Learning Model on Social Media' (Study \#13) first implemented preprocessing steps to balance the dataset, enhancing the model's ability to learn from both classes equally. After addressing the class imbalance, the study then used the F1 score and related metrics to evaluate model performance, ensuring a more accurate and fair assessment. These examples indicate that researchers are increasingly aware of the class imbalance issue and are employing various approaches to address it effectively.

Conversely, some studies primarily relied on accuracy without addressing class imbalance issues. For example, Studies \#2, \#10, and \#24 reported high accuracy but did not mention techniques to mitigate the effects of class imbalance.

In the context of depression detection, addressing class imbalance is essential for achieving reliable model evaluation. When instances of the non-depressed class significantly outnumber those of the depressed class, the resulting imbalance can skew model outcomes if not properly managed. Two primary strategies are commonly employed to mitigate this issue: the use of evaluation metrics that accommodate class imbalance and data preprocessing techniques, such as resampling and reweighting. \citeA{japkowicz2002imbalance} emphasize that metrics like recall, precision, and F1 score offer a more nuanced evaluation by accounting for both positive and negative classes, thus reducing potential bias. Additionally, data preprocessing methods like reweighting or resampling adjust the dataset to provide a balanced exposure to both classes, enhancing model training on imbalanced data.

While some studies utilized both strategies, demonstrating a thorough approach to handling imbalance, others employed just one—either through preferred evaluation metrics or data balancing. Even when only one strategy is adopted, it can still reduce potential bias to some extent. However, solely relying on accuracy introduces a significant risk of bias, as it often leads the model to favor the majority class, thereby failing to identify depressed individuals accurately. \citeA{chawla2004imbalanced} highlight that this reliance on accuracy alone can lead to misleading conclusions in imbalanced datasets, as it does not accurately reflect the model’s ability to detect minority class instances.

Out of the 47 studies analyzed, approximately 35 employed preferred metrics such as F1 score, precision, recall, or AUROC, recognizing their importance in evaluating models on imbalanced datasets. Seven studies explicitly mentioned preprocessing steps like resampling to mitigate class imbalance, even when using accuracy as an evaluation metric. However, several studies relied mainly on accuracy without addressing class imbalance, potentially introducing bias into their evaluations.

In conclusion, while a significant number of studies have adopted appropriate evaluation metrics and techniques to address class imbalance, there remains a need for broader implementation of these practices. Incorporating balanced metrics and addressing class imbalance is essential for reliable and valid model evaluations in depression detection research. As \citeA{fernandez2018imbalanced} recommended, employing these strategies enhances the robustness of machine learning models in domains characterized by imbalanced datasets.

\subsection{Reporting: Transparency and Completeness}
Transparency and completeness in reporting are fundamental to the integrity and reproducibility of scientific research. In our examination of the 47 studies, we assessed the extent to which they transparently reported their methodologies, findings, and limitations. Notably, all studies (100\%) included a limitation section, indicating an overall acknowledgment of the importance of addressing potential shortcomings. However, the depth and specificity of these disclosures varied significantly across the studies.

While every study mentioned limitations, not all of them fully recognized or disclosed all critical methodological issues that could impact their findings. For instance, as highlighted in our earlier assessments, approximately 23\% of the studies (11 out of 47) did not properly partition their data or failed to report their data partitioning methods adequately (Studies \#2, \#5, \#9, \#12, \#20, \#27, \#31, and \#37). Despite this, only a few of these studies explicitly acknowledged the potential biases introduced by improper data partitioning in their sections of limitations. This suggests that while researchers are generally aware of the necessity to report limitations, there is a gap in fully understanding or disclosing specific methodological shortcomings, such as data partitioning, which is crucial for model generalizability and validity.

Similarly, in the context of hyperparameter tuning, approximately 43\% of the studies did not report or properly tune hyperparameters across all models used (e.g., Studies \#1, \#2, \#4, \#5, \#12, \#14, \#17, \#19, \#20, \#24, \#27, \#29, \#30, \#32, \#34, \#35, \#37, \#38, \#42, \#44, and \#46). Only a few acknowledged this limitation in their reports. This lack of comprehensive reporting on hyperparameter tuning can lead to biased model comparisons and affect the reproducibility of the studies.

Incomplete or non-transparent reporting can introduce significant bias and limit the reproducibility and applicability of research findings. When critical methodological details are omitted or underreported, it hinders the ability of other researchers to replicate studies or to understand the context in which the results are valid. For instance, failing to disclose improper data partitioning can lead to overestimation of model performance due to overfitting \cite{bishop2006ml}. Models evaluated on training data or without appropriate validation may appear to perform well, but this performance may not generalize to new, unseen data. This oversight can mislead stakeholders about the efficacy of the models and affect subsequent research or practical applications that build upon these findings.

Similarly, not reporting on hyperparameter tuning practices can result in unfair comparisons between models and misinterpretations of their relative performances \cite{claesen2015hyperparameter, zhang2025tutorialusingmachinelearning}. Models with optimized hyperparameters may outperform others not because they are inherently better but because they were given an optimization advantage. Without transparency in reporting these practices, readers cannot assess the fairness of the comparisons or replicate the optimization process.

In conclusion, while all 47 studies recognized the importance of reporting limitations, there remains a notable disparity in the thoroughness and transparency of their reporting. For the field to advance, transparent and comprehensive reporting of methodologies and limitations is essential. Future research should strive for complete disclosure of data collection, preprocessing, model development, hyperparameter tuning, and evaluation metrics. This includes acknowledging specific methodological limitations, such as data partitioning practices and sampling biases, and discussing how these limitations may impact results and generalizability. Such transparency will allow others to interpret findings accurately, replicate studies, and build upon prior work effectively.

\subsection{Summary of Findings and Implications for Future Research}
This systematic review evaluated biases throughout the entire lifecycle of machine learning and deep learning models for depression detection on social media. In sampling, biases arose from a predominant reliance on Twitter, English-language data, and specific geographic regions, limiting the representativeness of findings. Data preprocessing commonly showed inadequate handling of negations, which can skew sentiment analysis results. Model development was often compromised by inconsistent hyperparameter tuning and improper data partitioning, reducing model reliability and generalizability. Lastly, in model evaluation, an overreliance on accuracy without addressing class imbalance risked favoring majority class predictions, potentially misleading results. These findings highlight the importance of enhancing methodologies to bolster the validity and applicability of future research.

To address these biases, future research should improve practices across all stages of the machine learning lifecycle. Expanding data sources across multiple platforms, languages, and regions will help mitigate platform and language biases and improve representativeness. Standardizing data preprocessing, especially with explicit negation handling, and employing resampling and reweighting techniques will enhance sentiment analysis accuracy and balance datasets. Consistent hyperparameter tuning protocols are essential to ensure fair model comparisons and optimal performance. Lastly, prioritizing evaluation metrics like precision, recall, F1 score, and AUROC in imbalanced datasets, particularly for depression detection, will yield more accurate and insightful assessments. By implementing these improvements, future studies can achieve greater model robustness and generalizability, contributing to more effective mental health detection tools.

\section{Discussion}
The escalating prevalence of mental health conditions, particularly depression, poses a significant global health challenge. Social media platforms have emerged as rich data sources where individuals express their thoughts and emotions, offering a unique opportunity to detect mental health issues through advanced computational methods. Machine learning and deep learning models hold promise for analyzing this vast, unstructured data to identify patterns indicative of depression. This systematic review aimed to evaluate the effectiveness of these models in detecting depression on social media, focusing on identifying and analyzing biases throughout the ML lifecycle.

\subsection{Summary of Key Findings}
Our review uncovered several key biases and methodological challenges that impact the reliability and generalizability of machine learning and deep learning models in this domain. Sampling biases emerged due to a predominant reliance on specific social media platforms, particularly Twitter, which was used in 63.8\% of the studies. Additionally, most studies focused on English-language content and users from specific geographic regions, primarily the United States and Europe. These biases limit the representativeness of findings, as they do not capture the diversity of global social media users. In data preprocessing, many studies inadequately handled linguistic nuances, such as negations and sarcasm. Only about 23\% of the studies explicitly addressed the handling of negative words or negations, which are crucial for accurate sentiment analysis in depression detection.

Model development issues were also prominent. Inconsistent hyperparameter tuning practices were observed, with only 27.7\% of the studies properly tuning hyperparameters for all models. Moreover, approximately 17\% of the studies did not adequately partition their data into training, validation, and test sets. These practices can lead to overfitting, reducing the models’ ability to generalize to new data. Regarding model evaluation, many studies relied heavily on accuracy as the primary evaluation metric without addressing class imbalances inherent in depression detection datasets. While about 74.5\% of the studies used metrics suitable for imbalanced data, such as precision, recall, F1 score, and AUROC, others did not, potentially skewing the evaluation of model performance. Finally, despite all studies including a limitations section, transparency varied significantly, with critical methodological details like data partitioning methods and hyperparameter settings often underreported. This inconsistency hinders reproducibility and the ability to fully assess the validity of the findings.

\subsection{Strengths and Limitations of the Review}
This systematic review stands out for its comprehensive scope, examining biases across the entire ML lifecycle, from sampling to reporting, in depression detection on social media. By not limiting the analysis to specific aspects, the review offers a holistic view of how biases can influence model validity. Another strength is the structured methodological approach, adhering to established guidelines with a well-defined search strategy and clear inclusion criteria. Focusing on studies published after 2010, it reflects the latest advancements in ML and DL applications for mental health.

The use of established bias assessment tools, particularly PROBAST, adds rigor by systematically evaluating bias across key methodological domains. Additionally, the review’s detailed data extraction process facilitated a structured analysis, allowing for the identification of patterns and providing actionable recommendations, such as diversifying data sources and improving transparency.

However, the review also has limitations. Limited database coverage and the English-only restriction may exclude valuable insights from non-English research, potentially affecting the generalizability of the findings. The focus on recent studies (post-2010) might have overlooked earlier influential works, while heterogeneity in study designs hindered direct comparisons and precluded a quantitative meta-analysis. Moreover, publication bias could skew findings toward positive results, and excluding grey literature means emerging methodologies may not be fully captured. Lastly, while ethical considerations were acknowledged, a deeper exploration of issues like data privacy and informed consent is warranted.

These limitations suggest areas for improvement in future research, such as broadening database and language coverage, including grey literature, and conducting a meta-analysis where feasible. By addressing these areas, future studies can enhance the robustness of ML models for mental health detection and provide a more comprehensive, ethical, and globally relevant understanding of the field.

\subsection{Implications for Future Research}
To enhance the generalizability and applicability of machine learning and deep learning models in depression detection on social media, addressing identified biases is essential. First, diversifying data sources across multiple social media platforms and incorporating non-English languages and underrepresented regions will improve representativeness and generalizability. Improving sampling methods is crucial. Combining keyword-based sampling with random sampling techniques can help reduce selection bias and capture users who may not explicitly mention depression but exhibit relevant behaviors. In the data preprocessing step, researchers should standardize practices to explicitly handle linguistic nuances like negations and sarcasm, which are vital for accurate sentiment analysis. Additionally, applying resampling or reweighting techniques can help balance datasets, ensuring that both classes—particularly the minority depressive class—are adequately represented. Advanced natural language processing techniques that account for linguistic nuances, such as sarcasm and context-dependent meanings, should be employed.

Consistent and comprehensive hyperparameter tuning across all models is essential to ensure fair comparisons and optimize model performance. Proper data partitioning practices, including the use of validation and test sets, should be implemented to prevent overfitting and assess model generalizability. When evaluating models, researchers should prioritize metrics that account for class imbalance, such as precision, recall, F1 score, and AUROC. These metrics provide a more balanced assessment of model performance and are more informative in the context of detecting depression, where the minority class is of primary interest.

\subsection{Concluding Remarks}
This systematic review highlights significant methodological limitations in current research on detecting depression through social media analysis using machine learning and deep learning models. Addressing these limitations is critical to developing more accurate, reliable, and generalizable models that can effectively identify individuals at risk of depression. Future research should focus on diversifying data sources, improving sampling methods, enhancing data preprocessing and model development practices, and employing appropriate evaluation metrics to ensure balanced and meaningful assessments.

By advancing these methodological approaches, researchers can contribute to the advancement of mental health detection tools that are ethically sound and effective across diverse populations and platforms. Such advancements hold the potential to facilitate early intervention strategies, ultimately improving mental health outcomes on a global scale.

\section{Appendix: Reviewed Studies on Machine Learning Models for Depression Detection on Social Media}
\renewcommand{\thetable}{A\arabic{table}} % Prefix table numbers with 'A'
\setcounter{table}{0} % Reset table counter if needed

\small
\begin{longtable}{c|p{10cm}|p{4cm}}
\caption{Reviewed Studies on Machine Learning Models for Depression Detection on Social Media} \\
\hline
\textbf{Index} & \textbf{Title of the Paper} & \textbf{Reference} \\
\hline
\endfirsthead
\hline
\textbf{Index} & \textbf{Title of the Paper} & \textbf{Reference} \\
\hline
\endhead
\hline
\multicolumn{3}{r}{\textit{Continued on next page}} \\
\endfoot
\hline
\endlastfoot
\#1 & Machine learning of language use on Twitter reveals weak and non-specific predictions & \citeA{kelley2022anxiety} \\
\hline
\#2 & Supervised machine learning models for depression sentiment analysis & \citeA{obagbuwa2023sentiment} \\
\hline
\#3 & A textual-based featuring approach for depression detection using machine learning classifiers and social media texts & \citeA{chiong2021depression} \\
\hline
\#4 & Emotional Distress During COVID-19 by Mental Health Conditions and Economic Vulnerability: Retrospective Analysis of Survey-Linked Twitter Data With a Semisupervised Machine Learning Algorithm & \citeA{ueda2023emotional} \\
\hline
\#5 & Depression and Anxiety on Twitter During the COVID-19 Stay-At-Home Period in 7 Major U.S. Cities & \citeA{levanti2023depression} \\
\hline
\#6 & Depression detection for Twitter users using sentiment analysis in English and Arabic tweets & \citeA{helmy2024depression} \\
\hline
\#7 & Deep Learning With Anaphora Resolution for the Detection of Tweeters With Depression: Algorithm Development and Validation Study & \citeA{wongkoblap2021depression} \\
\hline
\#8 & Sentiments about Mental Health on Twitter-Before and during the COVID-19 Pandemic & \citeA{beier2023sentiments} \\
\hline
\#9 & Hype or hope? Ketamine for the treatment of depression: results from the application of deep learning to Twitter posts from 2010 to 2023 & \citeA{ng2024ketamine} \\
\hline
\#10 & Quantifying depression-related language on social media during the COVID-19 pandemic & \citeA{davis2020depression} \\
\hline
\#11 & Predicting state-level suicide fatalities in the United States with realtime data and machine learning & \citeA{patel2023suicide} \\
\hline
\#12 & Investigating Social Media to Evaluate Emergency Medicine Physicians' Emotional Well-being During COVID-19 & \citeA{agarwal2023emotional} \\
\hline
\#13 & Explainable depression detection with multi-aspect features using a hybrid deep learning model on social media & \citeA{zogan2022depression} \\
\hline
\#14 & Big data analytics on social networks for real-time depression detection & \citeA{angsung2022bigdata} \\
\hline
\#15 & An optimized deep learning approach for suicide detection through Arabic tweets & \citeA{baghdadi2022suicide} \\
\hline
\#16 & COVID-19 sentiment analysis via deep learning during the rise of novel cases & \citeA{chandra2021covid} \\
\hline
\#17 & A Scalable Framework to Detect Personal Health Mentions on Twitter & \citeA{yin2015healthmentions} \\
\hline
\#18 & An automatic diagnostic network using skew-robust adversarial discriminative domain adaptation to evaluate the severity of depression & \citeA{sun2019depression} \\
\hline
\#19 & Twitter: a good place to detect health conditions & \citeA{prieto2014twitter} \\
\hline
\#20 & Consumer perceptions of telehealth for mental health or substance abuse: A Twitter-based topic modeling analysis & \citeA{baird2022telehealth} \\
\hline
\#21 & Depressive Emotion Detection and Behavior Analysis of Men Who Have Sex With Men via Social Media & \citeA{li2020men} \\
\hline
\#22 & Areas of Interest and Social Consideration of Antidepressants on English Tweets: A Natural Language Processing Classification Study & \citeA{deanta2022antidepressants} \\
\hline
\#23 & An Optimistic Firefly Algorithm-Based Deep Learning Approach for Sentiment Analysis of COVID-19 Tweets & \citeA{swapnarekha2023sentiment} \\
\hline
\#24 & How Do You \#relax When You're \#stressed? A Content Analysis and Infodemiology Study of Stress-Related Tweets & \citeA{doan2017relax} \\
\hline
\#25 & Psychological Disorder Identifying Method Based on Emotion Perception over Social Networks & \citeA{zhou2019disorder} \\
\hline
\#26 & Momentary Depressive Feeling Detection Using X (Formerly Twitter) Data: Contextual Language Approach & \citeA{jamali2023momentary} \\
\hline
\#27 & A Proposed Sentiment Analysis Deep Learning Algorithm for Analyzing COVID-19 Tweets & \citeA{kaur2021tweets} \\
\hline
\#28 & A machine learning approach predicts future risk to suicidal ideation from social media data & \citeA{roy2020suicide} \\
\hline
\#29 & Studying expressions of loneliness in individuals using Twitter: an observational study & \citeA{guntuku2019loneliness} \\
\hline
\#30 & Automatic Profiles Collection from Twitter Users with Depressive Symptoms & \citeA{wongkoblap2023profiles} \\
\hline
\#31 & Semi-Supervised Approach to Monitoring Clinical Depressive Symptoms in Social Media & \citeA{yazdavar2017depressive} \\
\hline
\#32 & Synthesis of Affective Expressions and Artificial Intelligence to Discover Mental Distress in Online Community & \citeA{singh2022affective} \\
\hline
\#33 & DAC Stacking: A Deep Learning Ensemble to Classify Anxiety, Depression, and Their Comorbidity from Reddit Texts & \citeA{borba2022dac} \\
\hline
\#34 & A textual-based featuring approach for depression detection using machine learning classifiers and social media texts & \citeA{chiong2021depression} \\
\hline
\#35 & Detection of Suicidality Among Opioid Users on Reddit: Machine Learning-Based Approach & \citeA{yao2020suicidality} \\
\hline
\#36 & Social Media Markers to Identify Fathers at Risk of Postpartum Depression: A Machine Learning Approach & \citeA{shatte2020postpartum} \\
\hline
\#37 & Social Media Mining for Postpartum Depression Prediction & \citeA{trifan2020postpartum} \\
\hline
\#38 & Social Media Discussions Predict Mental Health Consultations on College Campuses & \citeA{saha2022social} \\
\hline
\#39 & Enabling Early Health Care Intervention by Detecting Depression in Users of Web-Based Forums using Language Models: Longitudinal Analysis and Evaluation & \citeA{owen2023early} \\
\hline
\#40 & Natural Language Processing Reveals Vulnerable Mental Health Support Groups and Heightened Health Anxiety on Reddit During COVID-19: Observational Study & \citeA{low2020vulnerable} \\
\hline
\#41 & A deep learning model for detecting mental illness from user content on social media & \citeA{kim2020media} \\
\hline
\#42 & Classification of Helpful Comments on Online Suicide Watch Forums & \citeA{kavuluru2016suicide} \\
\hline
\#43 & Predicting future mental illness from social media: A big-data approach & \citeA{thorstad2019future} \\
\hline
\#44 & Depression detection from social network data using machine learning techniques & \citeA{islam2018detection} \\
\hline
\#45 & Deep neural networks detect suicide risk from textual Facebook posts & \citeA{ophir2020suicide} \\
\hline
\#46 & Leveraging Social Media to Predict COVID-19-Induced Disruptions to Mental Well-Being Among University Students: Modeling Study & \citeA{swain2024social} \\
\hline
\#47 & Building a profile of subjective well-being for social media users & \citeA{chen2017wellbeing} \\
\hline
\label{tabA1:studies}
\end{longtable}

\section*{Acknowledgments}
The authors thank Dr. Jin (Veronica) Liu (ORCID: 0000-0001-5922-6643) for her valuable comments and insightful feedback on the development of this manuscript. Her input has contributed to improving the clarity and overall presentation of the work.

\section*{Funding}
No funding was received for conducting this study.

\section*{Conflicts of Interest / Competing Interests}
The authors declare that they have no conflicts of interest or competing interests.

\section*{Ethics Approval}
Not applicable.

\section*{Consent to Participate}
Not applicable.

\section*{Consent for Publication}
All authors have read and agreed to the published version of the manuscript.

\section*{Availability of Data and Materials}
The reviewed titles, authors, and publication years of the included studies have been provided in Table A.1. Detailed information on each reviewed paper is hosted on GitHub: \texttt{https://github.com/odile1999/Systematic-Review-Machine-Learning-on-\linebreak Depression}.

\section*{Code Availability}
Not applicable.

\section*{Authors' Contributions}
Project administration: Y.T. and Y.C.; Conceptualization: Y.T. and Y.C.; Methodology: Y.T., J.D., and Y.C.; Investigation: Y.T., Y.C., J.D., Z.W., Y.Z., X.S., and Y.L.; Formal Analysis: Y.T., Y.C., J.D., Z.W., Y.Z., X.S., and Y.L.; Writing - Original Draft: Y.T., Y.C., and J.D.; Writing - Review and Editing: Y.C., Z.W., Y.Z., X.S., and Y.L.

\bibliographystyle{apacite}
\bibliography{short_bib}
\end{document}